\documentclass[]{fairmeta}

\usepackage{amsmath}
\usepackage{tikz}
\usetikzlibrary{shapes,arrows.meta,positioning,fit,calc}
\usepackage[table]{xcolor}
\usepackage{tcolorbox}
\usepackage{amsmath}
\usepackage{graphicx} 

\usepackage{booktabs}
\usepackage{multirow} 

\usepackage[utf8]{inputenc} 
\usepackage[T1]{fontenc}    
\usepackage{hyperref}       
\usepackage{url}            
\usepackage{booktabs}       
\usepackage{amsfonts}       
\usepackage{nicefrac}       
\usepackage{microtype}      
\usepackage{xcolor}         
\usepackage{graphicx}
\usepackage{float}

\usepackage{times}
\usepackage{latexsym}

\usepackage[T1]{fontenc}

\usepackage[utf8]{inputenc}

\usepackage{microtype}

\usepackage{inconsolata}

\usepackage{graphicx}

\usepackage{colortbl}
\usepackage{makecell}
\usepackage{lipsum}
\usepackage{array}

\definecolor{blue}{RGB}{109,157,211} 

\definecolor{red}{RGB}{226,130,154} 

\definecolor{gray}{RGB}{117,118,111}    
\definecolor{green}{RGB}{107,158,101}   

\usepackage{caption} 
\usepackage{float}
\title{TRACE: A Framework for Analyzing and Enhancing Stepwise Reasoning in Vision-Language Models}

\author{Shima Imani}
\author{Seungwhan Moon}
\author{Lambert Mathias}
\author{Lu Zhang}
\author{Babak Damavandi}

\affiliation{Meta Reality Lab}
\abstract{Reliable mathematical and scientific reasoning remains an open challenge for large vision-language models (VLMs). Standard final-answer evaluation often masks reasoning errors, allowing silent failures to persist. To address this gap, we introduce \textbf{TRACE}, a framework for \textit{Transparent Reasoning And Consistency Evaluation} that diagnoses reasoning trajectories rather than only end results. At its core, TRACE leverages \textbf{Auxiliary Reasoning Sets (ARS)}, compact sub-question–answer pairs that decompose complex problems, evaluate intermediate steps through consistency-based metrics, and expose failures overlooked by standard evaluation. Our experiments show that consistency across ARS correlates with final-answer correctness and helps pinpoint the reasoning steps where failures arise, offering actionable signals for model improvement. Furthermore, TRACE defines confidence regions that distinguish reliable from unreliable reasoning paths, supporting effective filtering, debugging, and model refinement. }

\date{\today}
\correspondence{First Author at \email{shimaimani@meta.com}}

\begin{document}

\maketitle

\section{Introduction}

\begin{figure*}[t]
    \centering
    \includegraphics[width=1\textwidth]{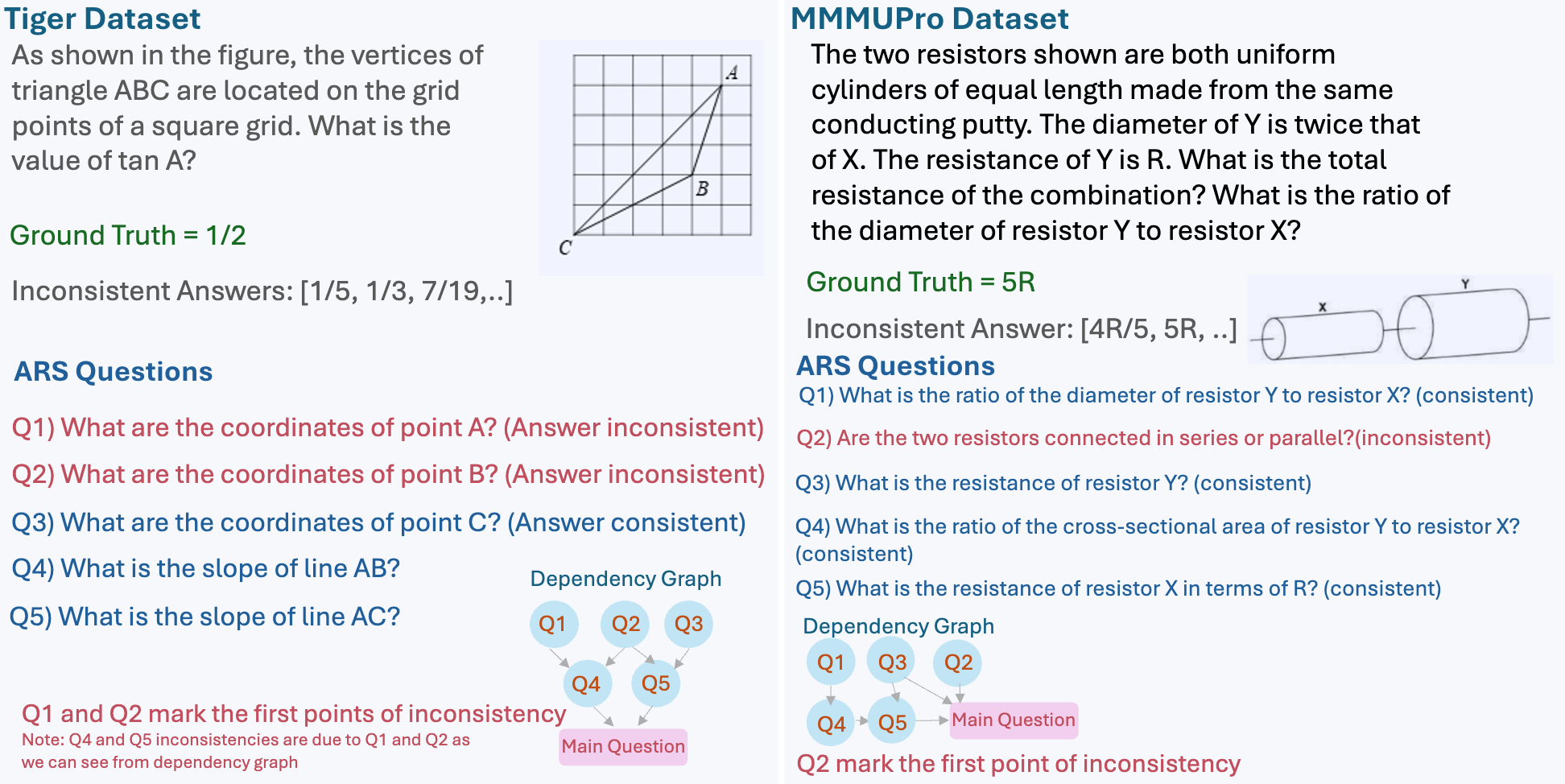}
    \caption{
    Illustration of the TRACE framework. 
    (\textbf{Left}) A geometry question from TIGER dataset \cite{virl39k} with its ARS decomposition: Q1–Q3 extract coordinates, Q4-Q5 compute slopes, and the final question computes $\tan A$. Nodes highlighted in \textcolor{red}{red} indicate inconsistent answers across model generations, allowing us to localize the source of error. 
    (\textbf{Right}) A physics problem from the MMMUPro dataset \cite{yue2024mmmu} with its ARS decomposition. Nodes highlighted in \textcolor{red}{red} indicate inconsistent answers across model generations, helping to pinpoint where the reasoning fails.
    }
    \label{fig:teaser}
\end{figure*}

Evaluating large vision–language models (VLMs) has predominantly focused on final-answer correctness. However, this metric is often insufficient and misleading. When a model produces an incorrect answer, it remains unclear \textit{where} in the reasoning process the failure occurred or how errors propagated through the computational graph. Conversely, a correct final answer does not guarantee a coherent reasoning process; models can reach correct outcomes through flawed or inconsistent intermediate steps, masking conceptual misunderstandings or accidental self-corrections. These limitations raise critical questions: {\itshape How can we pinpoint failures in multi-step, multimodal reasoning? What patterns produce “silent errors,” where incorrect intermediate steps still yield correct answers? And to what extent does final-answer accuracy truly reflect reasoning ability versus chance or memorization?}

To address these challenges, we introduce \textbf{TRACE} (\textit{Transparent Reasoning and Consistency Evaluation}), a framework that enhances diagnostic evaluation of VLM reasoning. TRACE systematically decomposes complex multimodal tasks into \textbf{Auxiliary Reasoning Sets (ARS)}: interpretable sub-questions with structured dependencies. By assessing consistency across ARS, TRACE precisely localizes reasoning failures and exposes error propagation. Unlike final-answer-centric evaluation, this decomposition reveals \textit{how consistently} a model reasons across sub-questions and which steps exhibit the most variation across reasoning paths, highlighting potentially unstable or ambiguous reasoning steps. In addition, TRACE maps ARS dependencies into a reasoning graph to identify the \textit{First Failure Step (FFS)}. As illustrated in Figure~\ref{fig:teaser}, this allows tracing inconsistencies across intermediate reasoning steps, revealing which sub-questions or dependencies exhibit the most variation, even when the final model answer does not fully reflect these differences.

\paragraph{Our contributions are:}
\begin{enumerate}
    \item We propose \textbf{TRACE}, a diagnostic framework for transparent multimodal reasoning, which decomposes complex tasks into Auxiliary Reasoning Sets (ARS) with a dependency-aware evaluation protocol for tracking error propagation.
    \item We introduce novel consistency metrics, including path mean consistency, global mean consistency, and the diagnostic \textit{First Failure Step (FFS)}, which reliably localize reasoning failures overlooked by traditional final-answer metrics.
    \item We construct a benchmark of 3.7k ARS question–answer pairs across 630 reasoning paths, enabling evaluation at both intermediate and final-answer levels.
    \item Experiments demonstrating TRACE uncovers reasoning errors missed by standard final-answer evaluation, enhancing interpretability, robustness, and evaluation quality.
\end{enumerate}

\section{Related Work}
\label{sec:related}
\paragraph{Multimodal reasoning benchmarks.}
Benchmarks like ScienceQA~\cite{lu2022learn}, MathVista~\cite{lu2023mathvista}, MMMU~\cite{yue2024mmmu}, and MathVision~\cite{wang2024mathvision} evaluate VLMs on STEM tasks involving text and diagrams. While they reveal progress and limitations, they mainly assess final-answer accuracy, which cannot distinguish genuine reasoning from lucky guesses.
\paragraph{Intermediate reasoning and process supervision.}
Methods such as chain-of-thought prompting~\cite{wei2022cot}, self-consistency~\cite{wang2022sc}, and process supervision (e.g., GSM8K~\cite{cobbe2021gsm8k}) improve performance and interpretability by incorporating intermediate steps. However, they often rely on natural language rationales, which are hard to evaluate automatically, or still depend on final-answer supervision, limiting error diagnosis in intermediate steps.
\paragraph{Problem decomposition and program-aided reasoning.}
Approaches like Program-of-Thoughts~\cite{chen2022pot}, PAL~\cite{gao2022pal}, and verifier-based pipelines~\cite{lightman2023lets} decompose reasoning into symbolic programs or use self-correction. While effective for arithmetic and robustness, they lack principled ways to diagnose failures at specific reasoning steps, especially in multimodal contexts.

\paragraph{Verifying and improving reasoning chains.}  
Prior work has examined the reliability of Chain-of-Thought (CoT) reasoning, showing that models often produce unfaithful traces~\cite{arcuschin2503chain, zhao2025chain, lanham2023measuring}. Studies highlight systematic biases where models justify contradictory answers or generate superficially coherent arguments~\cite{arcuschin2503chain}, and use interventions such as editing or truncating CoTs to test whether intermediate steps affect final answers~\cite{lanham2023measuring}. TRACE builds on these ideas by decomposing problems into Auxiliary Reasoning Sets, measuring consistency across intermediate steps, and identifying confidence regions to detect unreliable reasoning paths, offering actionable guidance for debugging and improving multimodal reasoning.

\section{Methodology}
\label{sec:methodology}

Building on these insights, we introduce \textbf{TRACE} a framework that structures and scrutinizes the intermediate reasoning steps of multimodal models. TRACE moves beyond conventional black-box mapping, where a model directly maps $\text{VLM(Question, Image)}$ to a final answer. Instead, it introduces an intermediate \textbf{Auxiliary Reasoning Set (ARS)}, 
\[
\mathcal{S} = \{(q_i, a_i)\}_{i=1}^n,
\]
consisting of sub-question–answer pairs that explicitly capture intermediate reasoning steps. An ARS is defined such that each sub-question $q_i$ paired with an answer $a_i$ satisfies three properties:
\begin{itemize}
    \item \textbf{Completeness:} The sub-questions collectively provide all information needed to solve the problem, with no redundancy. The ARS also extracts any required information from the image, so that the main question can be answered without directly referencing the figure.
    \item \textbf{Independence:} Each sub-question depends only on raw inputs or explicitly specified predecessors.
    \item \textbf{Soundness:} Sub-questions are answerable, non-overlapping, and do not leak the final answer.
\end{itemize}

The model then uses this structured set to produce its final answer, making the reasoning process transparent and enabling fine-grained evaluation of reasoning behavior.

\textbf{Example.} Consider a geometry problem illustrated in Figure~\ref{fig:teaser}, where the vertices of triangle ABC lie on a square grid and the task is to compute $\tan A$. The ARS decomposes the problem into sub-questions such as identifying the coordinates of points A, B, and C, and computing the slopes of lines AB and AC. This decomposition allows TRACE to analyze the model’s reasoning process, track how intermediate answers propagate, and provide a structured view of the entire solution pathway.

\textbf{Construction.} ARS are generated using two complementary strategies:
\begin{itemize}
    \item \textbf{Exploration:} Given the original question and a specialized prompt, the model generates diverse sub-questions along with their dependencies.  
    \item \textbf{Exploitation:} Sub-questions are generated from candidate reasoning chains in two steps:
    \begin{enumerate}
        \item \textit{Step 1:} Given the question and image, the model produces a step-by-step reasoning answer.  
        \item \textit{Step 2:} Using the original question, the generated reasoning answer, and a prompt, the model generates sub-questions and their dependencies, ensuring the ARS captures the critical steps necessary to solve the problem.  
    \end{enumerate}
\end{itemize}

\textbf{Reasoning Graph.} Each sub-question is annotated with metadata specifying its dependencies on other questions, text, or image inputs. Together, these dependencies form a directed acyclic reasoning graph (DAG) that encodes the structure of the reasoning process and defines the order in which the model answers sub-questions. Intermediate answers propagate forward through the DAG, and the final answer is computed based on these intermediate results. This structured execution allows TRACE to pinpoint exactly which reasoning steps lead to errors, enabling transparent analysis of the model's reasoning process.

\begin{table*}[ht]
\centering
\small
\begin{tabular}{p{2.6cm} p{0.8cm} ccc}
\toprule
\textbf{Model} & \textbf{Dataset} & \textbf{Path Mean Consistency} & \textbf{Path Z-score Consistency} & \textbf{Final Correctness} \\
\midrule
\multirow{2}{=}{{Llama-4-Maverick-17B-128E-Instruct}} & \multirow{2}{=}{TIGER} 
    & 0.79 & 3.83 & Incorrect \\
 & & 0.92 & 5.83 & Correct \\
\midrule
\multirow{2}{=}{{GPT-4.1}} & \multirow{2}{=}{MMMUPro} 
    & 0.787 & 3.77 & Incorrect \\
 & & 0.907 & 5.73 & Correct \\
\midrule
\multirow{2}{=}{{Qwen2.5-VL-72B-Instruct}} & \multirow{2}{=}{MMMUPro} 
    & 0.772 & 3.98 & Incorrect \\
 & & 0.853 & 4.88 & Correct \\
\midrule
\multirow{2}{=}{{Llama-4-Maverick-17B-128E-Instruct}} & \multirow{2}{=}{MMMUPro} 
    & 0.806 & 4.38 & Incorrect \\
 & & 0.903 & 5.62 & Correct \\
\bottomrule
\end{tabular}
\caption{Comparison of models on path consistency metrics and final correctness across datasets. Each row pair shows results split by correctness, demonstrating systematically higher stability for correct reasoning paths.}
\label{tab:consistency-results}
\end{table*}

\subsection{Evaluation Metrics}
TRACE evaluates reasoning performance at two complementary levels. First, \textbf{consistency-based metrics} measure stability: 
the \textbf{path-level} captures agreement within individual reasoning trajectories, while the \textbf{global-level} captures agreement across all paths for the same question. 
Second, the \textbf{First Failure Step (FFS)} identifies the earliest sub-question where a reasoning path diverges from consensus or shows inconsistency, providing a diagnostic signal for how local errors propagate to final outcomes. 
Together, these metrics offer a holistic view of model reliability, capturing both intermediate-step coherence and final-answer correctness.

\subsection{Definitions and Notation}

Let a question be decomposed into an Auxiliary Reasoning Set (ARS):
\[
\mathcal{S} = \{(q_i, a_i)\}_{i=1}^n,
\]
where each $q_i$ is a sub-question and $a_i$ is the corresponding answer. 
For each question, we sample $K$ reasoning paths, each independently answering all sub-questions and producing a final answer. We define:
\begin{itemize}
    \item $P_j$: The $j$-th reasoning path.
    \item $A_{i,j}$: Answer to sub-question $q_i$ in $P_j$.
    \item $C_{i,j}$: Number of paths that gave the same answer to sub-question $q_i$ as path $P_j$.

\end{itemize}

We define the following consistency metrics:

\begin{description}
\item[Path Mean Consistency (PMC):] measures the average agreement across all sub-questions in the path:
    \[
    \text{PMC}_j = \frac{1}{n} \sum_{i=1}^n C_{i,j}.
    \]

\item[Path Deviation Consistency:] Standard deviation of agreement scores across sub-questions in $P_j$:
\begin{equation}
\small
\text{PDC}_j = \sqrt{\tfrac{1}{n} \sum_{i=1}^n \big(C_{i,j} - \text{PMC}_j\big)^2}.
\end{equation}

\item[Path Z-score Consistency:] Normalized path consistency:
\begin{equation}
\small
\text{PZC}_j = \log \left(\frac{\sum_{i=1}^n C_{i,j} - \text{PMC}_j}{\text{PDC}_j}\right).
\end{equation}

\item[Global Mean Consistency:] Average agreement across all sub-questions and all paths for a given question:
\begin{equation}
\small
\text{GMC} = \frac{1}{nK} \sum_{i=1}^n \sum_{j=1}^K C_{i,j}.
\end{equation}

\item[Consistency Gap:] Difference between a path’s mean consistency and the global average:
\begin{equation}
\small
\text{CG}_j = \text{PMC}_j - \text{GMC}.
\end{equation}\label{def:consistency_gap}

\end{description}

\paragraph{First Failure Step (FFS):}\label{def:ffs}
For each question, we define the FFS as the earliest sub-question in a reasoning path that contributes to an \emph{incorrect} final answer (or deviates from the majority outcome when no gold answer is available). Let a reasoning path $P_j$ have sub-questions $\{q_1, \dots, q_n\}$ with answers $\{A_{1,j}, \dots, A_{n,j}\}$. Denote the consensus answer for sub-question $q_i$ across consistent paths as $\hat{A}_{maj}$. Then the FFS for path $P_j$ is

\begin{align}
\text{FFS}(P_j) = \min \Bigl\{ \, 
i \in \{1, \dots, n\} \;\big|\; A_{i,j} \neq \hat{A}_{\text{maj}}, \nonumber\\
\text{and path $P_j$ yields a wrong final answer} \, \Bigr\}.
\end{align}

The FFS identifies the first local deviation that propagates into a final failure, providing a diagnostic signal for reasoning breakdowns.

Together, these metrics provide a fine-grained and interpretable evaluation of reasoning behavior, enabling analysis of both the reliability and robustness of VLM reasoning at intermediate steps and at the final-answer level.

\section{Experiments and Results}

\subsection{Dataset Overview}
We evaluate TRACE on two challenging STEM benchmarks that require multi-step reasoning over both visual and symbolic information:

\paragraph{MMMUPro (Vision).}  
This multimodal STEM benchmark includes a wide range of problems with figures, diagrams, and symbolic equations, requiring reasoning over both visual and textual information. For our experiments, we focus on three subjects: Mathematics, Physics, and Chemistry, to evaluate TRACE across diverse STEM domains \cite{yue2024mmmu}.

\paragraph{TIGER (Verifiable Subset).} We focus on a curated subset of 500 questions with ground-truth, verifiable solutions. Each problem is structured to support multi-step reasoning, making it ideal for assessing not only final-answer correctness but also the consistency of intermediate reasoning steps \cite{virl39k}.

\subsection{ARS Generation}
For both datasets, TRACE generates ARS for each problem using \texttt{Llama-4-Maverick-17B-128E-Instruct}, following the exploration and exploitation strategies described in Section~\ref{sec:methodology}. For each ARS set, multiple reasoning paths are generated across different models by sampling with temperature between 0 and 0.5 and nucleus sampling with $top‑p = 0.9$. During dataset preparation, we manually inspected 5\% of the generated ARS and iteratively refined prompts to improve quality. To ensure reliability, ARS whose performance fell below baseline models were filtered out. Further details on ARS quality and statistics are provided in Appendix~\ref{arsquality}.

We further compared the effectiveness of exploration-based and exploitation-based ARS. While exploration outperformed on some problems and exploitation on others, their overall accuracy was broadly similar across the datasets (Table~\ref{table:comparison_explore_exploit}; full results are reported in the Appendix).

\subsection{Results and Analysis}

\begin{figure*}[t]
    \centering
    \includegraphics[width=0.91\textwidth]{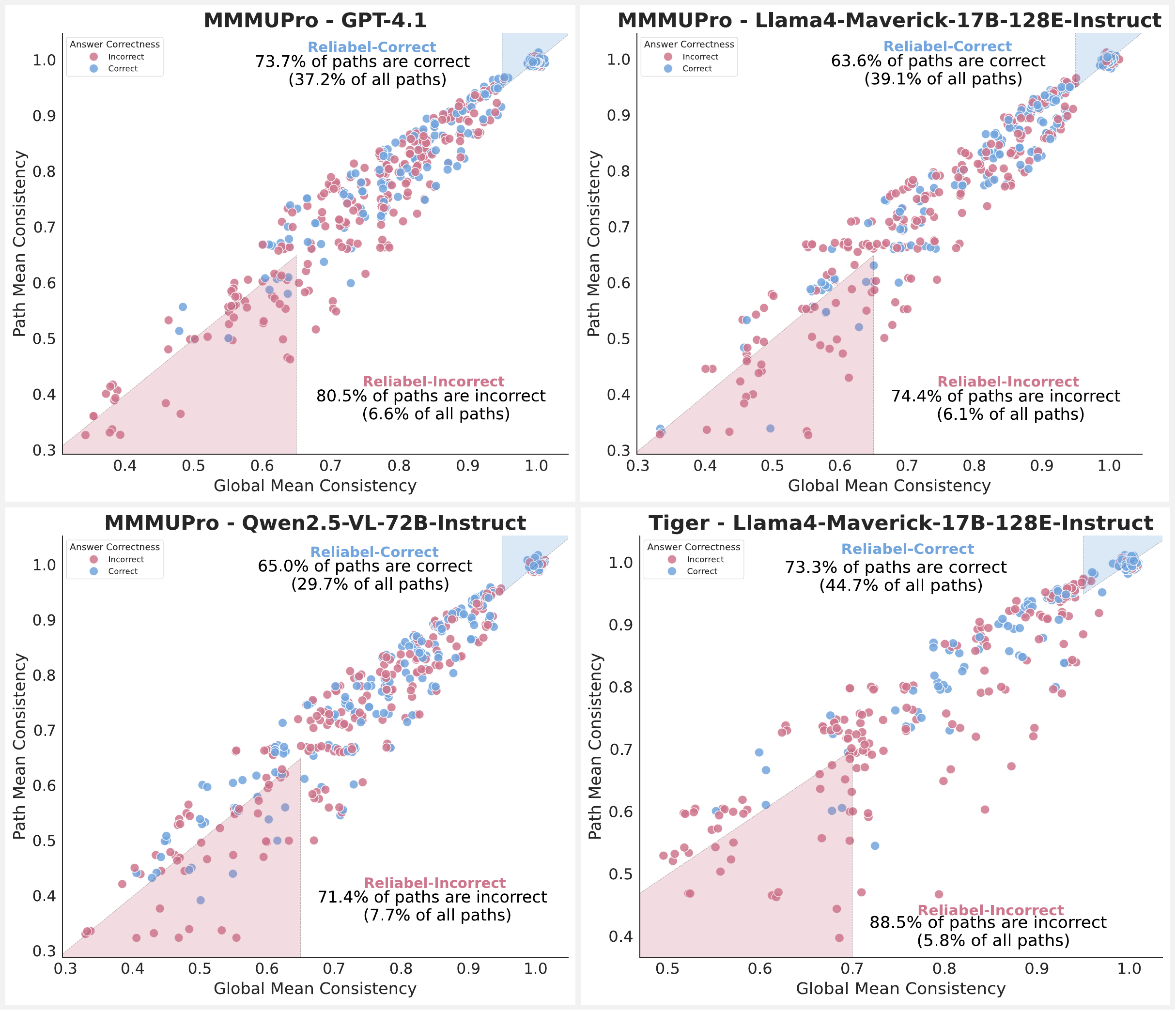}
    \caption{Relationship between path mean consistency (PMC) and global mean consistency (GMC) across models on MMMUPro. Blue regions indicate \textcolor{blue}{Reliable-Correct}, red regions indicate \textcolor{red}{Reliable-Incorrect}, and all remaining paths fall into the \textcolor{gray}{Uncertain} region.
}
    \label{fig:scatter_consistency}
\end{figure*}

\subsubsection{Correctness vs.\ Path Consistency}
We first analyze the direct relationship between path consistency metrics and final answer correctness. 
Table~\ref{tab:consistency-results} compares reasoning paths that produce correct versus incorrect outcomes. 
Across all models and datasets, correct paths exhibit substantially higher stability across sub-questions. 
In particular, they achieve both higher \textit{Path Mean Consistency (PMC)} and higher \textit{Path Z-score Consistency (PZC)} than incorrect paths. This analysis establishes that sub-question consistency carries predictive value: more consistent trajectories are substantially more likely to yield correct outcomes. 
While this demonstrates useful correlation, it remains a descriptive view. To unlock stronger predictive power, we next analyze path consistency relative to the question-level baseline.

\subsubsection{Consistency Gap Analysis}
\begin{figure}[H]
    \centering
    \includegraphics[width=0.5\textwidth]{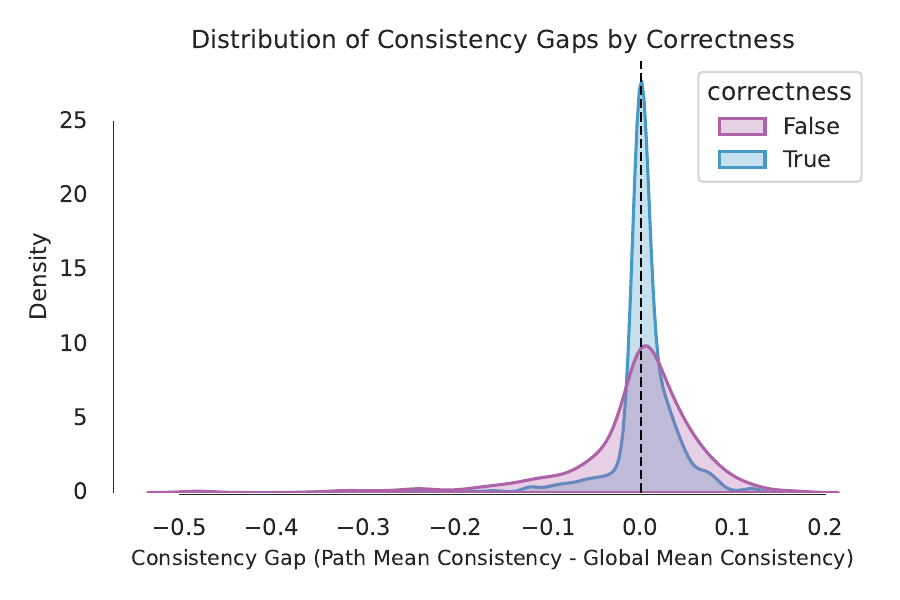}
    \caption{Distribution of consistency gaps by correctness. Correct paths skew right of zero (above-average consistency), while incorrect paths are broadly dispersed and often below zero.}
    \label{fig:consistency_gap}
\end{figure}

We introduce the \textbf{consistency gap} (Def.~\ref{def:consistency_gap}), defined as the deviation between a path’s PMC and the Global Mean Consistency (GMC) of all paths for that question. 
This relative measure situates each trajectory within its competitive context, providing a more nuanced lens on reliability. Figure~\ref{fig:consistency_gap} shows the distribution of consistency gaps on the TIGER dataset. 
Correct paths are concentrated around slightly positive values, meaning their consistency exceeds the question-level average.

Incorrect paths, by contrast, are widely dispersed and often fall below the mean, reflecting weaker internal stability. 
This trend generalizes across datasets and models, indicating that relative path consistency offers a principled basis for distinguishing reliable from unreliable reasoning.

\subsubsection{Confidence Region Definition}
\begin{figure*}[t]
    \centering
    \includegraphics[width=0.8\textwidth]{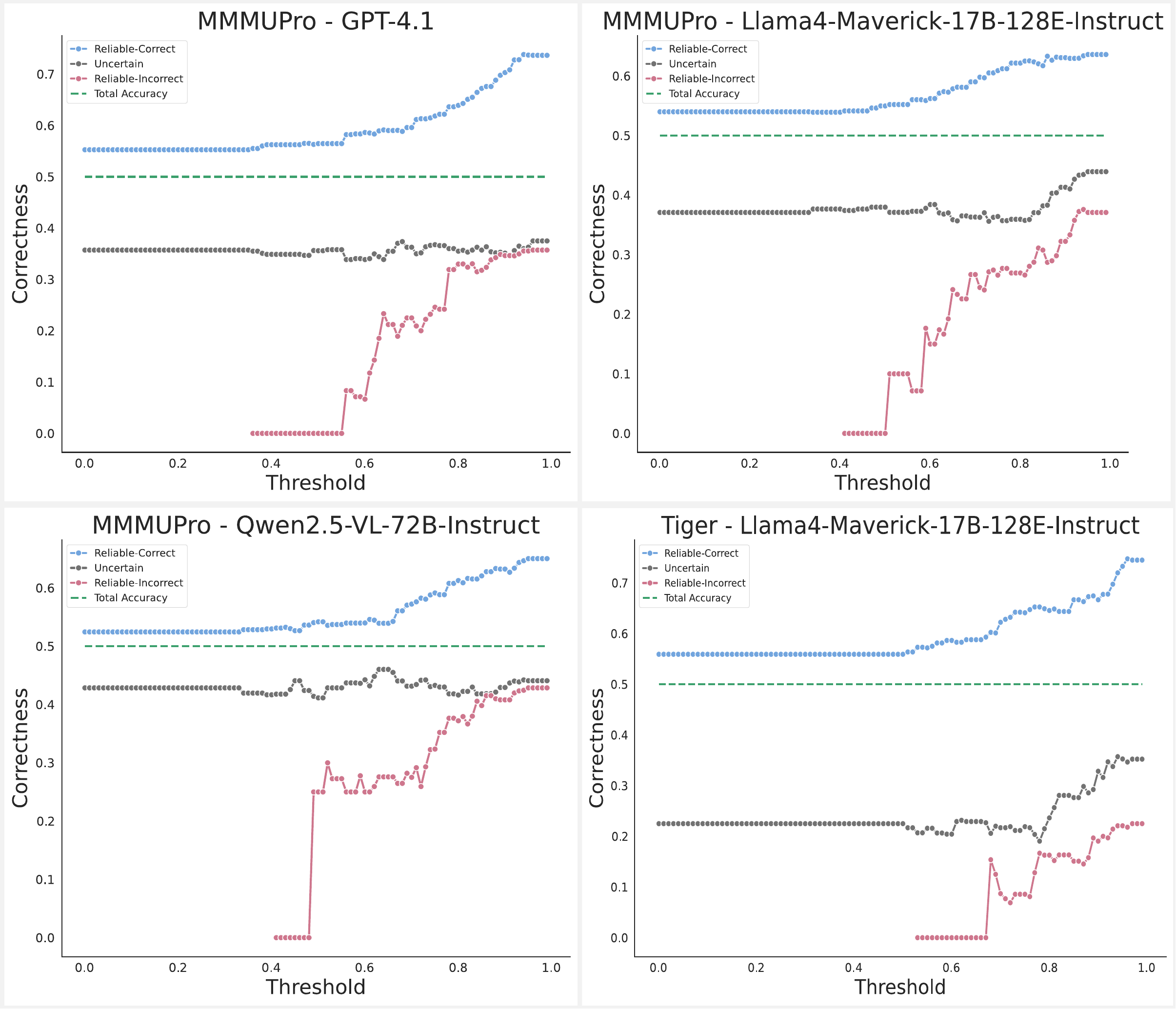}
    \caption{Accuracy of reasoning paths under different confidence regions as a function of threshold $t$. \textcolor{blue}{Reliable-Correct} and \textcolor{red}{Reliable-Incorrect} provide strong predictive signals. \textcolor{gray}{Uncertain} remains close to chance levels.}
    \label{fig:threshold_accuracy}
\end{figure*}
Building on the consistency gap signal, we define three interpretable regions to categorize reasoning paths:

\begin{enumerate}
    \item \textbf{\textcolor{blue}{Reliable-Correct Region (blue):}} Paths where the GMC exceeds a threshold $t$ and the PMC is greater than or equal to the GMC. These paths are strongly associated with correct final answers. Intuitively, they are not only consistent at the global question level but also internally more coherent than peer paths, reflecting robust reasoning.
    
    \item \textbf{\textcolor{red}{Reliable-Incorrect Region (red):}} Paths where the GMC is below a threshold $t$ and the PMC is lower than the GMC. These paths are strongly associated with incorrect answers, exhibiting both low agreement and poor relative stability. They serve as strong indicators of systematic unreliability.
    
    \item \textbf{\textcolor{gray}{Uncertain Region (gray):}} All remaining paths that do not fall into the Reliable-Correct or Reliable-Incorrect categories. These trajectories display mixed or intermediate consistency patterns, making correctness less predictable.
\end{enumerate}


Figure~\ref{fig:scatter_consistency} illustrates the relationship between Path Mean Consistency (PMC) and Global Mean Consistency (GMC) across models. The diagonal line $PMC = GMC$ serves as the boundary for relative consistency. The \textcolor{blue}{Reliable-Correct} region (above the diagonal) is dominated by correct paths (blue dots), while the \textcolor{red}{Reliable-Incorrect} region (below the diagonal) is heavily populated by incorrect paths (red dots). For instance, on the MMMU-Pro dataset, the GPT-4.1 model shows that approximately 73.7\% of paths in the Reliable-Correct region are correct, whereas around 80.5\% of paths in the Reliable-Incorrect region are incorrect. This distinct clustering demonstrates that the criteria $GMC \geq t$ and $PMC \geq GMC$ effectively identify paths in the \textcolor{blue}{Reliable-Correct} region, which are strongly predictive of correct answers. Conversely, paths in the \textcolor{red}{Reliable-Incorrect} region, which are highly predictive of errors, can be filtered or result in the model responding with “I don’t know” rather than producing an incorrect answer.

Figure~\ref{fig:threshold_accuracy} further examines how the classification of paths into Reliable-Correct, Reliable-Incorrect, and Uncertain regions changes as the Global Mean Consistency threshold ($t$) is varied. The results show that these regions remain predictive across thresholds: paths classified as Reliable-Correct at higher thresholds are largely correct, while paths classified as Reliable-Incorrect at lower thresholds are mostly incorrect. This behavior is also evident in Figure~\ref{fig:scatter_consistency}, where the triangular regions clearly capture the clustering of correct and incorrect paths.


\subsubsection{First Failure Step (FFS) Analysis}

Consistency-based metrics capture the overall stability of reasoning paths but do not reveal \emph{where} a trajectory first breaks down. To address this, we introduce the \textbf{First Failure Step (FFS)}, a diagnostic measure that identifies the earliest sub-question whose incorrect answer propagates to a wrong final outcome~\ref{def:ffs}. FFS enables fine-grained error analysis, pinpointing the exact step where reasoning fails and distinguishing isolated mistakes from systematic errors, making it a valuable tool for model evaluation and improvement of reasoning strategies. To illustrate this, consider the following example:



\begin{figure}[H]
    \centering
    \includegraphics[width=0.58\linewidth]{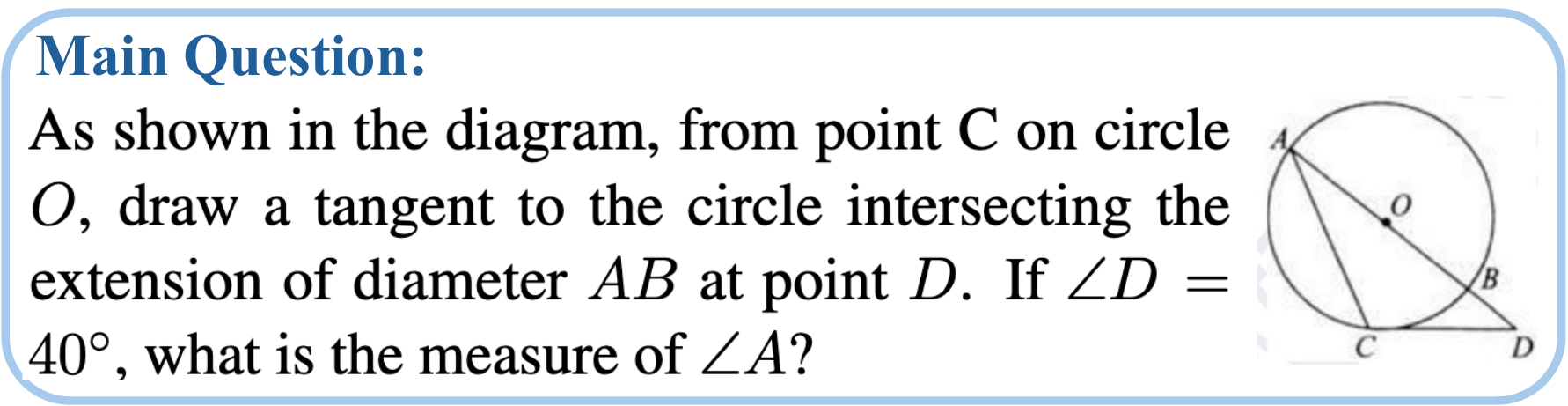}
    \label{fig:ffs-question-diagram}
\end{figure}

The ARS for the main question is as follows:

\begin{table}[H]
\small
\centering
\begin{tabular}{@{}p{0.45\columnwidth} p{0.45\columnwidth}@{}}
\toprule
\textbf{ARS Sub-question + Main Question} & \textbf{Path Answer / Majority Answer} \\
\midrule
Q1: What is the measure of $\angle D$? & 40 / \textbf{40} \\
Q2: Is $CD$ a tangent to circle $O$? & Yes / \textbf{Yes} \\
Q3: What is the measure of $\angle OCD$? & 90 / \textbf{90} \\
Q4: What is the measure of $\angle COD$? & 50 / \textbf{50} \\
\rowcolor{red!10}
Q5: What is the measure of $\angle AOC$? & \textcolor{red}{130} / \textbf{80} \textit{(First Incorrect)} \\
\arrayrulecolor{gray} 
\noalign{\vskip 2pt} 
\hline
\noalign{\vskip 2pt} 
\textbf{Main Question}: What is the measure of $\angle A$? & \textcolor{red}{65 (Incorrect)} / \textcolor{green}{\textbf{25 (Correct)}} \\
\arrayrulecolor{black} 
\bottomrule
\end{tabular}
\caption{Sub-questions and corresponding path answers vs. majority answers for an FFS example.}
\label{tab:ffs-example}
\end{table}

\begin{figure*}[t]
\centering
\includegraphics[width=0.85\textwidth]{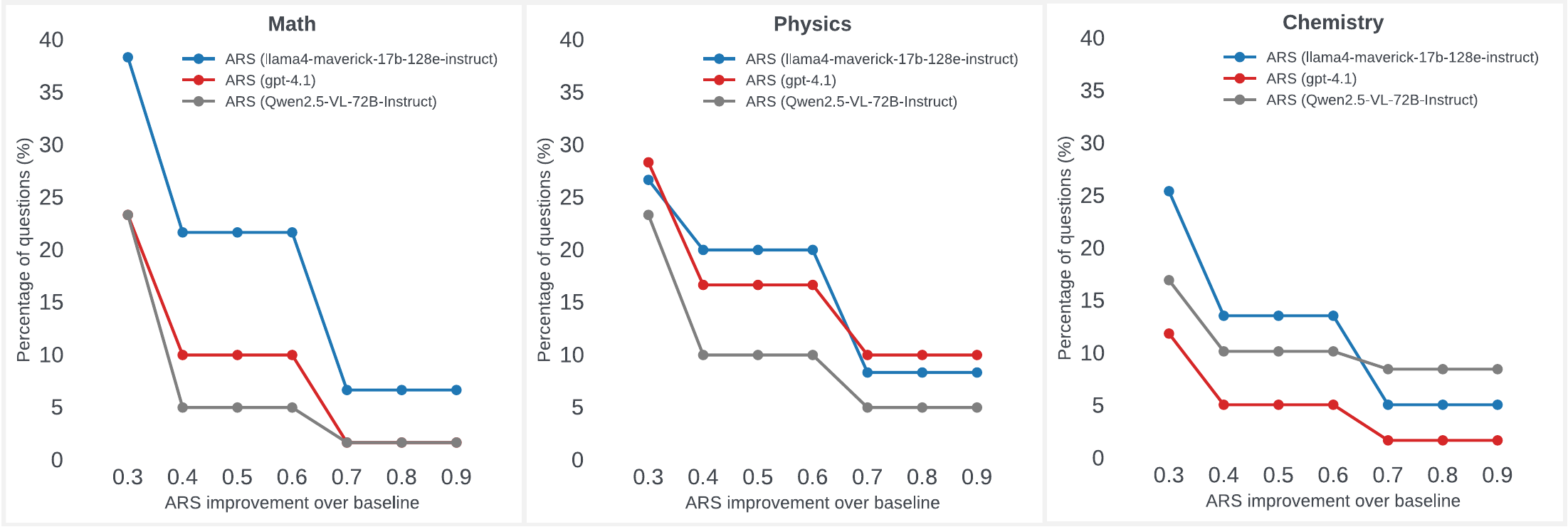}
\caption{ARS-guided improvement over the unstructured baseline across models on thenMMMUPro dataset. The x-axis shows the increase in final-answer accuracy, while the y-axis shows the fraction of questions achieving at least that improvement.}
\label{fig:ars_vs_baseline_subjects}
\end{figure*}

 This example illustrates the importance of the ARS sub-question \textit{``What is the measure of $\angle AOC$?''} in evaluating multi-step reasoning. Correctly answering this sub-question is crucial because it directly determines the main question, \(\angle A\), through the `Inscribed Angle Theorem': \(\angle A\) is an inscribed angle subtending the same arc \(\widehat{AC}\) as the central angle \(\angle AOC\), and therefore
\[
\angle A = \frac{1}{2} \angle AOC.
\] 
In Table~\ref{tab:ffs-example}, the model predicts \(\angle AOC = 130^\circ\) while the correct majority answer is \(80^\circ\), marking the \textit{First Failure Step (FFS)}. This demonstrates that even a single incorrect intermediate answer can propagate to the final prediction (\(\angle A = 65^\circ\) vs. correct \(25^\circ\)). By including such pivotal sub-questions in the ARS, TRACE can reliably localize reasoning failures, highlight critical dependencies, and provide actionable insights for debugging and improving VLM performance on multi-step tasks. Appendix~\ref{ffs-exampleIII} presents further examples of inconsistent reasoning patterns, extending beyond simple deviations from the majority.





\subsubsection{Subject-wise ARS Performance}

The primary purpose of ARS is to enhance interpretability by exposing intermediate reasoning dependencies, making it possible to trace which sub-steps contribute to correct or incorrect answers. Beyond interpretability, ARS also produces measurable gains in final-answer accuracy. Figure~\ref{fig:ars_vs_baseline_subjects} shows the fraction of questions achieving different levels of improvement across Math, Physics, and Chemistry for three models. Llama-4-Maverick-17B consistently benefits the most, with approximately 37\% of Math questions improving by at least 0.3, compared to about 23\% for GPT-4.1 and Qwen2.5-VL-72B. Even at the highest improvement threshold of 0.9, Llama-4-Maverick still improves around 7\% of Math questions, demonstrating ARS’s ability to correct particularly challenging errors.

The results also highlight the subject-specific value of structured reasoning: Math and Physics questions, which typically require multi-step derivations and precise intermediate calculations, gain the most from ARS, whereas Chemistry shows smaller but notable improvements. These findings suggest that ARS helps models maintain consistency and correctness throughout complex reasoning trajectories, rather than relying on shortcut reasoning that might produce a correct final answer by chance.
This demonstrates the potential of ARS as a training signal. By providing high-quality, stepwise reasoning paths, ARS can guide models to learn structured problem-solving strategies (see Appendix~\ref{sec:subject-wise-performance} for detailed examples, statistics, and further analyses).

\section{Conclusion and Future Work}

We introduce \textbf{TRACE} (\textit{Transparent Reasoning And Consistency Evaluation}), a framework for analyzing reasoning in VLMs. TRACE decomposes complex multimodal problems into \textbf{Auxiliary Reasoning Sets (ARS)}, offering a structured view of intermediate reasoning steps. Our results demonstrate that consistency across ARS correlates with final-answer correctness, providing a principled way to identify reliable and unreliable reasoning paths. Additionally, \textbf{First Failure Step (FFS)} detects the earliest point of deviation in reasoning, uncovering errors that standard evaluation metrics often overlook.

 Future work includes integrating these consistency signals into model training, for example by using reasoning traces as structured rewards in reinforcement learning frameworks such as GRPO to reduce reward hacking. We also plan to extend TRACE beyond STEM domains to broader multimodal reasoning tasks and real-world scenarios. 
 These directions aim to enhance the interpretability and robustness of multimodal reasoning models.

\clearpage
\newpage
\bibliographystyle{assets/plainnat}
\bibliography{paper}

\clearpage
\newpage
\beginappendix

\paragraph{FFS Example II} 
\label{appendix:ffs-example-2}

Consider another case where the model deviates from the majority at an early sub-question. Here, the final answer is incorrect, with the first error occurring at Q2, highlighting how early mistakes propagate and why identifying FFS is crucial for debugging multi-step reasoning.

\begin{figure}[h]  
\textbf{Main Question:}

In the Cartesian coordinate system $xOy$, a circle centered at the origin $O$ passes through the point $A(13,0)$. The line $y = kx - 3k + 4$ intersects the circle $\odot O$ at points $B$ and $C$. What is the minimum length of the chord $BC$?

\vspace{0.5em}
\centering
\includegraphics[width=0.25\linewidth]{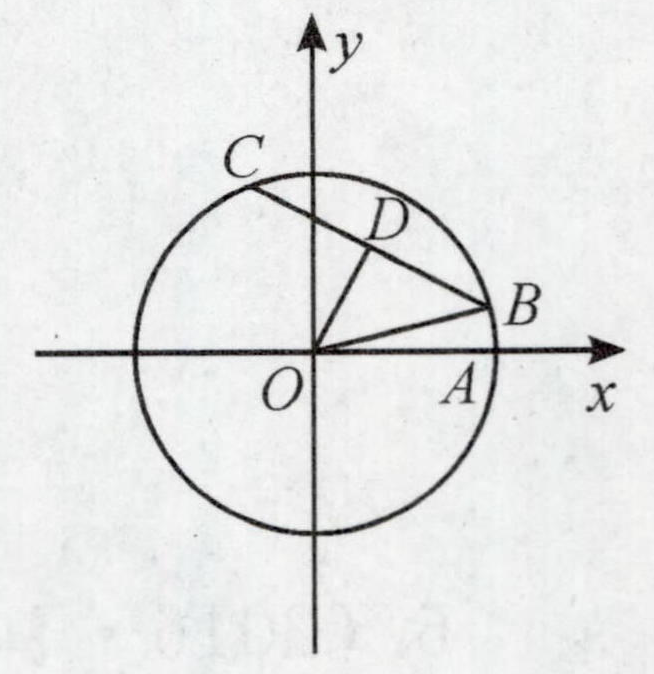}
\label{fig:ffs-question-diagram-2}
\end{figure}

\begin{table}[H]
\small
\centering
\begin{tabular}{@{}p{0.45\columnwidth} p{0.45\columnwidth}@{}}
\toprule
\textbf{ARS Sub-question + Main Question} & \textbf{Path Answer / Majority Answer} \\
\midrule
Q1: What is the radius of the circle centered at $O$? & 13 / \textbf{13} \\
\rowcolor{red!10}
Q2: What is the distance from $O$ to the line $y = kx - 3k + 4$? & \textcolor{red}{4} / \textbf{5} \textit{(First Incorrect)} \\
\arrayrulecolor{gray} 
\noalign{\vskip 2pt} 
\hline
\noalign{\vskip 2pt} 
\textbf{Main Question}: What is the minimum length of chord $BC$? & $6\sqrt{17}$ / \textbf{24} \\
\bottomrule
\end{tabular}
\caption{Sub-questions and corresponding path answers vs. majority answers for the second FFS example.}
\label{tab:ffs-example-2}
\end{table}

\begin{tcolorbox}[
    colback=gray!5,
    colframe=gray!70,
    boxrule=0.5pt,
    arc=2pt,
    left=6pt, right=6pt, top=6pt, bottom=6pt,
    title=Why the Distance to the Line is Critical
]
\small
The correct computation of the distance from $O$ to the line $y = kx - 3k + 4$ (Q2) is essential, as it directly determines the chord length $BC$.

\medskip
\textbf{Geometric Relationship:}
\begin{itemize}
    \item The perpendicular distance $d$ from $O$ to the line sets the position of the chord within the circle.
    \item For a circle of radius $r$, the chord length is:
    \[
    |BC| = 2\sqrt{r^2 - d^2}
    \]
    \item With $r = 13$ and $d = 5$, the minimum chord length is:
    \[
    |BC| = 2\sqrt{13^2 - 5^2} = 2\sqrt{169 - 25} = 24
    \]
\end{itemize}

\end{tcolorbox}

\paragraph{Illustrative Example III} \label{ffs-exampleIII} 
Consider a reasoning path from the MMMUPro dataset where the final answer is incorrect. FFS identifies the first sub-question in the path where the model's response deviates from the expected majority. In this example, the GPT-4.1 model provides inconsistent answers to Q3 across iterations—once identifying \textit{``$B$ and $C$ are the foci''} and another time \textit{``$A$ and $C$''}. This inconsistency marks Q3 as the initial point of reasoning failure.
This illustrates how ARS dependencies help isolate faulty visual reasoning steps.

\begin{figure}[h]  
\textbf{Main Question:}

The elliptical orbit of a planet around the Sun is shown in the diagram. Which of the following statements is true?

\centering
\includegraphics[width=0.25\linewidth]{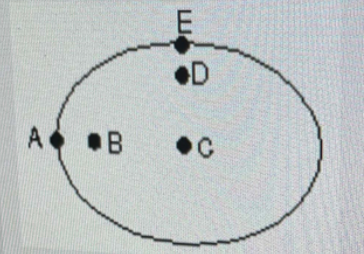}
\label{fig:ffs-mmupro-diagram}
\end{figure}

\begin{table}[h]
\small
\centering
\begin{tabular}{@{}p{0.45\columnwidth} p{0.45\columnwidth}@{}}
\toprule
\textbf{ARS Sub-question + Main Question} & \textbf{Path Answer I/ Path Answer II} \\
\midrule
Q1: What is the shape of the planet's orbit as shown in the diagram? & ellipse / \textbf{ellipse} \\
Q2: How many focal points does the elliptical orbit have in the diagram? & 2 / \textbf{2} \\
\rowcolor{red!10}
Q3: Which labeled points (A, B, C, D, E) are located at the foci of the ellipse? & \textcolor{red}{B and C} / \textcolor{red}{A and C} \textit{(First Incorrect)} \\
Q4: Is the Sun located at one of the foci of the elliptical orbit according to the diagram? & Yes / \textbf{Yes} \\
Q5: Is the eccentricity of the orbit equal to zero? & No / \textbf{No} \\
Q6: Is the eccentricity of the orbit equal to one? & No / \textbf{No} \\
Q7: Is the eccentricity of the orbit greater than one? & No / \textbf{No} \\
Q8: Is the eccentricity of the orbit less than zero? & No / \textbf{No} \\
\arrayrulecolor{gray} 
\noalign{\vskip 2pt} 
\hline
\noalign{\vskip 2pt} 
\textbf{Main Question}: Which of the following statements is true? & the sun might be at point C / \textbf{the sun might be at point A or C} (both incorrect) \\
\bottomrule
\end{tabular}
\caption{Sub-questions and corresponding path answers vs. majority answers for a MMMUPro orbit example. The first reasoning failure occurs at Q3.}
\label{tab:ffs-mmupro}
\end{table}

\begin{tcolorbox}[
    colback=gray!5,
    colframe=gray!70,
    boxrule=0.5pt,
    arc=2pt,
    left=4pt, right=4pt, top=4pt, bottom=4pt,
    title=Why Identifying the Foci Matters
]
\footnotesize
The sub-question \textit{``Which labeled points are located at the foci of the ellipse?''} is pivotal for determining whether the Sun is correctly placed in the diagram. 

\medskip
\textbf{Geometric Relationship:}
\begin{itemize}\itemsep1pt
    \item An ellipse has exactly two foci.
    \item Kepler's First Law: A planet orbits the Sun in an ellipse with the Sun at one focus.
    \item Misidentifying the foci propagates to an incorrect placement of the Sun and therefore an incorrect final answer.
\end{itemize}

\end{tcolorbox}

\paragraph{Illustrative Example IV} 
\label{appendix:ffs-example-4}

This example presents a First Failure Step (FFS) in a geometric ratio computation. The model’s reasoning fails at Q3, which propagates to an incorrect final area for $\triangle ABC$.

\begin{figure}[h]  
\textbf{Main Question:}

In $\triangle ABC$, $E$ is the midpoint of $AC$, and $D$ lies on $BC$ such that $BD:CD = 2:3$. Lines $AD$ and $BE$ intersect at $O$. If $S_{\triangle AOE} - S_{\triangle BOD} = 1$, find the area of $\triangle ABC$.

\vspace{0.5em}
\centering
\includegraphics[width=0.25\linewidth]{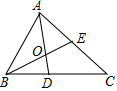}
\label{fig:ffs-question-diagram-4}
\end{figure}

\begin{table}[h]
\small
\centering
\begin{tabular}{@{}p{0.45\columnwidth} p{0.45\columnwidth}@{}}
\toprule
\textbf{ARS Sub-question + Main Question} & \textbf{Path Answer / Majority Answer} \\
\midrule
Q1: What is the ratio of $BD$ to $CD$? & $(2/3)$ / \textbf{$(2/3)$} \\
Q2: Is $E$ the midpoint of $AC$? & Yes / \textbf{Yes} \\
\rowcolor{red!10}
Q3: What is the ratio of areas $\triangle AOE : \triangle BOD$? & \textcolor{red}{4} / \textbf{$(5/4)$} \textit{(First Incorrect)} \\
Q5: What is the area of $\triangle ABE$? & 6 / \textbf{5} \\
Q6: What is the area of $\triangle ABD$? & $(56/9)$ / \textbf{4} \\
\arrayrulecolor{gray} 
\noalign{\vskip 2pt} 
\hline
\noalign{\vskip 2pt} 
\textbf{Main Question}: Area of $\triangle ABC$? & $(140/9)$ / \textbf{10} \\
\bottomrule
\end{tabular}
\caption{Sub-questions and path vs. majority answers}
\label{tab:ffs-example-4}
\end{table}

\paragraph{First Failure Step (FFS):}  
The model first fails at Q3, incorrectly computing the ratio of areas $\triangle AOE : \triangle BOD$ as $4$ instead of $5/4$. This miscalculation propagates to subsequent area computations, leading to an incorrect area for $\triangle ABC$.

\begin{tcolorbox}[
    colback=gray!5,
    colframe=gray!70,
    boxrule=0.5pt,
    arc=2pt,
    left=6pt, right=6pt, top=6pt, bottom=6pt,
    title=Why Area Ratios Determine the Final Answer
]
\small
The ratio $\dfrac{S_{\triangle AOE}}{S_{\triangle BOD}}$ links the condition
$S_{\triangle AOE} - S_{\triangle BOD} = 1$ to the absolute area of $\triangle ABC$.

\medskip
\textbf{Geometric Relationship:}
\begin{itemize}
    \item Intersection of cevians $AD$ and $BE$ defines proportional sub-triangles.
    \item Correct ratio ($5/4$) allows solving for $S_{\triangle AOE}$ and $S_{\triangle BOD}$.
    \item Summing sub-triangle areas scales to $S_{\triangle ABC} = 10$.
\end{itemize}

\textbf{ARS Dependency Justification:}  
Q3 directly governs the chain of reasoning. An incorrect ratio leads to errors in all subsequent area computations.
\end{tcolorbox}

\paragraph{Illustrative Example V} 
\label{appendix:ffs-example-5}

This example illustrates a First Failure Step (FFS) in coordinate reasoning. The model first fails at Q5, leading to an incorrect final reflection point $B'$.

\begin{figure}[h]  
\textbf{Main Question:}

In rectangle $OABC$, $OA$ and $OC$ lie on the $x$- and $y$-axes. $B=(3,2)$, points $D$ and $E$ lie on $AB$ and $BC$ with $BD = BE = 1$. Folding $\triangle BDE$ along $DE$ maps $B$ to $B'$. Find the coordinates of $B'$.

\vspace{0.5em}
\centering
\includegraphics[width=0.25\linewidth]{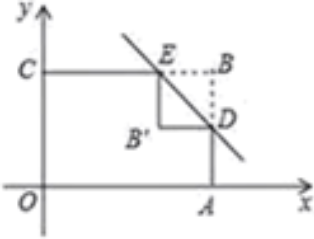}
\label{fig:ffs-question-diagram-5}
\end{figure}

\begin{table}[h]
\small
\centering
\begin{tabular}{@{}p{0.45\columnwidth} p{0.45\columnwidth}@{}}
\toprule
\textbf{ARS Sub-question + Main Question} & \textbf{Path Answer / Majority Answer} \\
\midrule
Q1: Coordinates of $B$? & $(3,2)$ / \textbf{$(3,2)$} \\
Q2: Length of $BD$? & $1$ / 1 \\
Q3: Length of $BE$? & $1$ / 1 \\
Q4: Coordinates of $D$? & $(3,1)$ / \textbf{$(3,1)$} \\
\rowcolor{red!10}
Q5: Coordinates of $E$? & \textcolor{red}{$(3,1)$} / \textbf{$(2,2)$} \textit{(First Incorrect)} \\
Q6: Slope of $DE$? & Undefined / \textbf{$-1$} \\
Q7: Equation of line $DE$? & $x=3$ / \textbf{$y=-x+4$} \\
\arrayrulecolor{gray} 
\noalign{\vskip 2pt} 
\hline
\noalign{\vskip 2pt} 
\textbf{Main Question}: Coordinates of $B'$? & $(3,0)$ / \textbf{$(2,1)$} \\
\bottomrule
\end{tabular}
\caption{Sub-questions and path vs. majority answers.}
\label{tab:ffs-example-5}
\end{table}

\paragraph{First Failure Step (FFS):}  
The model first fails at Q5 by incorrectly placing $E$ at $(3,1)$ instead of $(2,2)$. This error propagates to Q6 (slope), Q7 (line equation), and the final reflection, producing an incorrect $B'$. Identifying Q5 as the FFS reveals the cascading effect of a single misstep in coordinate placement.

\begin{tcolorbox}[
    colback=gray!5,
    colframe=gray!70,
    boxrule=0.5pt,
    arc=2pt,
    left=6pt, right=6pt, top=6pt, bottom=6pt,
    title=Why Correct Placement of $E$ is Critical
]
\small
Point $E$ lies on $BC$, one unit from $B(3,2)$. Correct placement gives $E=(2,2)$.

\medskip
\textbf{Consequences:}
\begin{itemize}
    \item Line $DE$ has slope $-1$ and equation $y=-x+4$.
    \item Folding across $DE$ maps $B(3,2)$ to $B'=(2,1)$.
    \item An incorrect coordinate at Q5 misaligns all subsequent computations.
\end{itemize}

\textbf{ARS Dependency Justification:}  
Q5 directly influences the line $DE$ and the final reflection. Local errors propagate through the reasoning chain, demonstrating how the FFS captures the earliest point of failure.
\end{tcolorbox}

\section{Appendix: ARS Generation Analysis}

\subsection{Exploration vs. Exploitation Performance}

As described in Section~\ref{sec:methodology}, TRACE generates Auxiliary Reasoning Sub-questions (ARS) using two complementary strategies: \textit{Exploration}, where diverse sub-questions are generated directly from the original problem, and \textit{Exploitation}, where sub-questions are derived from candidate reasoning chains. 

Table~\ref{table:comparison_explore_exploit} reports the accuracy of ARS generated by each strategy on MMMUPro before filtering low-quality ARS sets. We observe that, for most questions, exploration and exploitation achieve comparable accuracy, with some questions favoring one strategy over the other. This confirms that both approaches are viable for generating high-quality ARS, although the choice of strategy may influence performance for specific problem instances.

\begin{table}[h]
\centering
\small 
\setlength{\tabcolsep}{5pt} 
\renewcommand{\arraystretch}{0.9} 
\begin{tabular}{lcc}
\toprule
\textbf{Model} & \textbf{Exploitation} & \textbf{Exploration} \\
\midrule
Qwen2.5-VL-72B-Instruct & 0.385 & 0.375 \\
Maverick-17B-128E-Instruct & 0.534 & 0.537 \\
GPT-4.1 & 0.528 & 0.518 \\
\bottomrule
\end{tabular}
\caption{Exploitation vs.\ exploration ARS generation strategies on MMMUPro (before ARS filtering).}
\label{table:comparison_explore_exploit}
\end{table}

\subsection{Subject-wise ARS Performance Analysis}
\label{sec:subject-wise-performance}

The primary objective of the Auxiliary Reasoning Set (ARS) is to enhance interpretability and provide diagnostic signals. However, it is also informative to examine cases where ARS-guided answers improve final correctness over the unstructured baseline. Figure~\ref{fig:ars_vs_baseline_subjects} shows the fraction of questions exhibiting a given magnitude of improvement (x-axis) for three models across three core STEM subjects.

To illustrate the practical impact of structured ARS on model reasoning, we present a concrete example from MMMUPro. This case demonstrates how sequential sub-questions guide the model to improve the final answer results compared to the unstructured baseline. By inspecting the ARS-guided reasoning path, we can see how step-by-step clarification resolves ambiguities that the baseline model struggles with.

\begin{figure}[H]  
\textbf{Main Question:}

The diagram above depicts iron filings sprinkled around three permanent magnets. Pole R is the same pole as:

\centering
\includegraphics[width=0.25\linewidth]{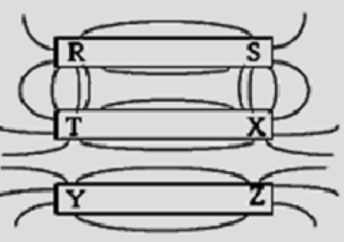}

\textbf{Options:} Y and Z, X and T, R and T, X and Y, T and Z, T and Y, R and Z, R and Y, X and Z

\label{fig:ars-mmupro-diagram}
\end{figure}

\begin{table}[H]
\small
\centering
\setlength{\tabcolsep}{5pt} 
\renewcommand{\arraystretch}{0.95} 
\begin{tabular}{@{}p{0.55\columnwidth} p{0.4\columnwidth}@{}}
\toprule
\textbf{ARS Sub-question + Main Question} & \textbf{ARS Path Answer} \\
\midrule
Q1: In the diagram, what is the direction of the magnetic field lines around pole R? & outward \\
Q2: In the diagram, what is the direction of the magnetic field lines around pole S? & into it \\
Q3: In the diagram, what is the direction of the magnetic field lines around pole T? & into it \\
Q4: In the diagram, what is the direction of the magnetic field lines around pole X? & outward \\
Q5: In the diagram, what is the direction of the magnetic field lines around pole Y? & outward \\
Q6: In the diagram, what is the direction of the magnetic field lines around pole Z? & towards Z \\
\arrayrulecolor{gray} 
\noalign{\vskip 2pt} 
\hline
\noalign{\vskip 2pt} 
\textbf{Main Question}: Pole R is the same pole as & \textcolor{green}{\textbf{X and Y (Correct)}} \\
\bottomrule
\end{tabular}
\caption{Example of ARS-guided reasoning on a magnetic field question. Only ARS answers are shown, illustrating how step-by-step sub-questions guide the model to the correct final answer.}
\label{tab:ars-mmupro}
\end{table}

\begin{tcolorbox}[colback=red!4!white, colframe=red!80!blue, title=Baseline Answer]
The model attempts to identify which pole is the same as R by analyzing the direction of magnetic field lines. It notes that field lines emerge from R and T and enter S and X, and from Y into Z. Based on this, the model concludes that T and Y are also North poles. Using this reasoning, it selects option F. However, this reasoning is incomplete and leads to an incorrect final answer. \textcolor{red}{Incorrect}
\end{tcolorbox}

\begin{table*}[h]
\centering
\small
\begin{tabular}{lccc}
\toprule
\textbf{Model} & $T=0.0$ & $T=0.2$ & $T=0.4$ \\
\midrule
Llama-4-Maverick-17B-128E-Instruct & 0.553 / 0.553 & 0.520 / 0.575 & 0.553 / 0.536 \\
GPT-4.1 & 0.514 / 0.413 & 0.492 / 0.453 & 0.441 / 0.475 \\
Qwen2.5-VL-72B-Instruct & 0.318 / 0.335 & 0.358 / 0.358 & 0.374 / 0.352 \\
\bottomrule
\end{tabular}
\caption{Accuracy for exploration / exploitation paths under varying sampling temperatures. These results illustrate the impact of stochasticity on ARS quality (Pre-Filtering)}
\label{tab:ars_temperature}
\end{table*}
\subsection{ARS Quality}\label{arsquality}
We evaluate the quality of Auxiliary Reasoning Sets (ARS) through automated filtering, manual inspection, and statistical analysis. Each ARS question is verified to ensure it does not leak information from the main question. Using the Llama-4-Maverick-17B-128E-Instruct model, each ARS question is checked for information leakage: sub-questions that essentially ask the same concept as the main question are removed. 

\paragraph{Baseline vs. ARS Accuracy (Pre-Filtering).}  
Table~\ref{tab:ars_vs_baseline} reports baseline model accuracies and ARS-guided reasoning accuracies before applying any filtering. This setting captures the raw contribution of ARS generation across datasets. Importantly, for the main question we do not provide image information directly; all visual content must be recovered within the ARS. In MMMUPro, where the question itself is embedded in the image, we first use the Llama-4-Maverick-17B-128E-Instruct model to extract the textual form of the question. The observed drop in pre-filtering accuracy arises because ARS sometimes fails to capture all necessary information from the image. To address this, we perform manual inspection and post-filtering to ensure completeness and reliability of ARS inputs.

\begin{table}[h]
\centering
\small
\setlength{\tabcolsep}{5pt}
\renewcommand{\arraystretch}{1.0}
\begin{tabular}{lccc}
\toprule
\textbf{Dataset} & \textbf{model} & \textbf{Baseline Acc (\%)} & \textbf{ARS Acc (\%)} \\
\midrule
MMMUPro & GPT4.1 & 54.3 & 48.2 \\
TIGER  & Maverick & 79.9 & 70.9 \\
\bottomrule
\end{tabular}
\caption{Baseline model accuracy versus ARS-guided reasoning accuracy (pre-filtering).}
\label{tab:ars_vs_baseline}
\end{table}

\begin{figure}[h]
    \centering
    \includegraphics[width=0.45\linewidth]{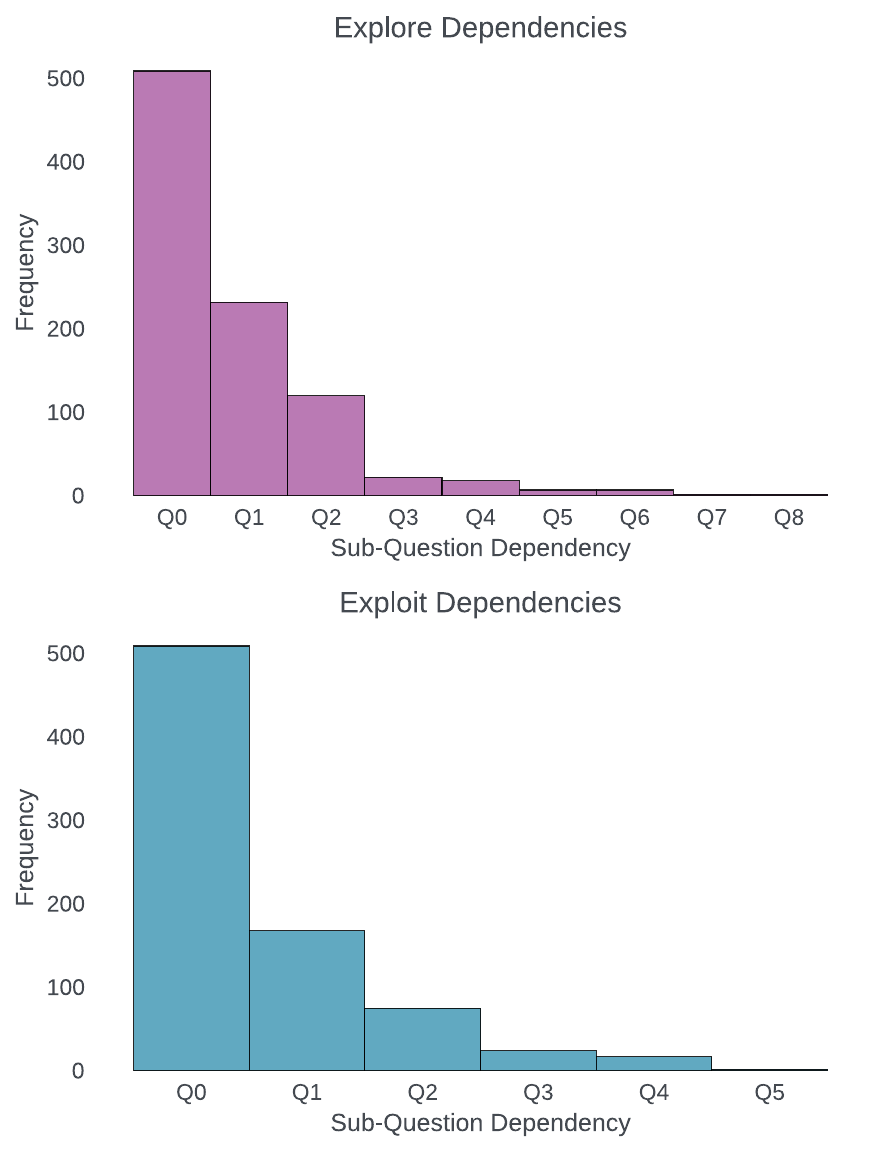}    \caption{Histograms of ARS sub-question dependencies for MMMUPro dataset for each strategy.}
    \label{fig:dependency_hist_mmmupro}
\end{figure}

\begin{figure}[h]
    \centering
    \includegraphics[width=0.45\linewidth]{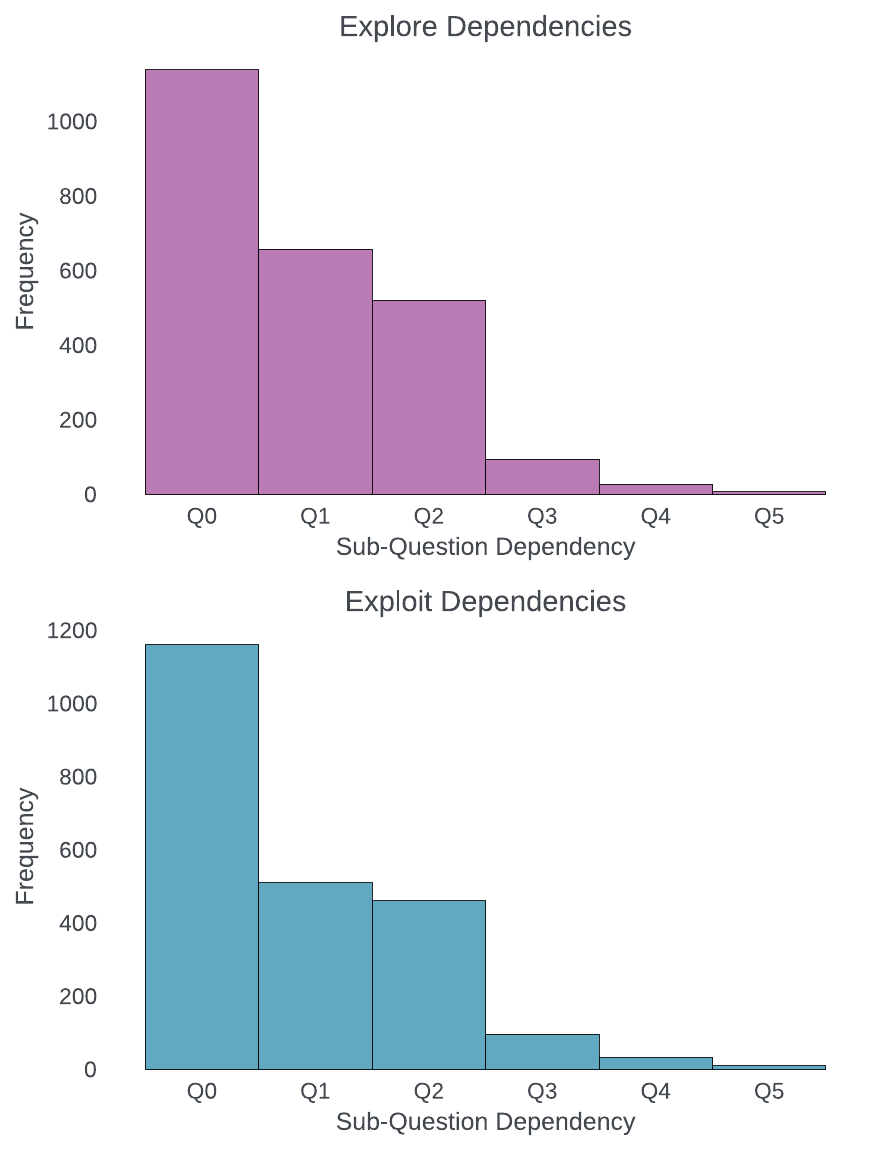}
    \caption{Histograms of ARS sub-question dependencies for TIGER dataset for each strategy.}
    \label{fig:dependency_hist_tiger}
\end{figure}

\paragraph{Manual Inspection.}  
We review 5\% of ARS, including those corresponding to low-baseline problems, to verify that even challenging questions generate meaningful and well-structured sub-questions. Insights from this manual inspection are used to iteratively refine the prompts, improving the overall quality and coverage of ARS. Additionally, any ARS whose average accuracy falls below the baseline is automatically filtered out, ensuring that only high-quality and informative sub-questions are retained.

\paragraph{ARS Statistics (Post-Filtering).}  
After filtering ARS sets for quality and leakage, key statistics for exploration and exploitation paths are reported in Table~\ref{tab:ars_exploitation} and \ref{tab:ars_exploration}. On average, MMMUPro requires about six sub-questions per problem, compared to roughly four for the TIGER dataset. This suggests that MMMUPro problems demand more reasoning steps. The higher complexity of MMMUPro is also reflected in accuracy: 54.3\% versus 79.9\% for TIGER (Table~\ref{tab:ars_vs_baseline}). Thus, ARS statistics not only capture structural properties of reasoning paths but also provide insights into dataset difficulty. Consistent with this, the average sub-question dependency is higher for MMMUPro than for TIGER, further indicating its greater reasoning complexity.
 Figures~\ref{fig:dependency_hist_tiger} and~\ref{fig:dependency_hist_mmmupro} present the distributions of sub-question dependencies for the TIGER and MMMUPro datasets, respectively. 
The dependency graph illustrates that the initial three questions function as the most frequent dependency nodes for all subsequent sub-questions, confirming the expected hierarchical structure of the task. This centralized structure highlights the sensitivity of the overall reasoning chain to the model's performance on these foundational steps, as errors introduced here are likely to propagate to dependent sub-questions.

\begin{table}[t]
\centering
\footnotesize
\setlength{\tabcolsep}{4pt}
\renewcommand{\arraystretch}{0.8}
\begin{tabular}{lccc}
\toprule
\multicolumn{4}{c}{\textbf{Exploration}} \\
\midrule
\textbf{Dataset} & Image & Total Qs & Dep. (Avg/Max) \\
\midrule
MMMUPro & 0.618 & 6.13 & 0.78 / 8 \\
TIGER   & 0.456 & 6.29 & 0.89 / 12 \\
\bottomrule
\end{tabular}
\caption{ARS statistics for the Exploration strategy: fraction of questions requiring images, total sub-questions per problem on average, and average/max sub-question dependencies.}
\label{tab:ars_exploration}
\end{table}

\vspace{0.5em} 

\begin{table}[t]
\centering
\footnotesize
\setlength{\tabcolsep}{4pt}
\renewcommand{\arraystretch}{0.8}
\begin{tabular}{lccc}
\toprule
\multicolumn{4}{c}{\textbf{Exploitation}} \\
\midrule
\textbf{Dataset} & Image & Total Qs & Dep. (Avg/Max) \\
\midrule
MMMUPro & 0.539 & 4.13 & 0.60 / 7 \\
TIGER   & 0.422 & 3.70 & 0.84 / 8 \\
\bottomrule
\end{tabular}
\caption{ARS statistics for the Exploitation strategy: fraction of questions requiring images, total sub-questions per problem on average, and average/max sub-question dependencies.}
\label{tab:ars_exploitation}
\end{table}



\subsection{Ablation: Temperature Effects on ARS Accuracy}  
We analyze the sensitivity of ARS quality to sampling temperature. For each ARS, multiple reasoning paths are generated with $T \in \{0.0, 0.2, 0.4\}$ using the Llama-4-Maverick-17B-128E-Instruct model. We then report the final-answer accuracy obtained with ARS, separately for exploration and exploitation strategies. Table~\ref{tab:ars_temperature} summarizes the results, where each cell reports accuracy for exploration (first value) and exploitation (second value).

\subsection{TRACE: ARS Generation Prompt}

To ensure reproducibility and clarity of our method, we provide the full prompt used by TRACE to generate Auxiliary Reasoning Sets (ARS). 
The prompt instructs the model to decompose a complex visual reasoning problem into minimal, structured sub-questions that are sufficient to solve the main question without accessing the image. 
For \textit{exploration}, only the original question is provided, whereas for \textit{exploitation}, the prompt additionally includes the candidate answer to guide sub-question generation (see Figure~\ref{generation_prompt}).

\begin{figure*}[t]
\centering
\begin{tcolorbox}[
  title={TRACE: ARS Generation Prompt},
  colback=gray!5,
  colframe=gray!70,
  boxrule=0.4pt,
  arc=2pt,
  fonttitle=\bfseries,
  left=4pt,
  right=4pt,
  top=4pt,
  bottom=4pt,
  width=\textwidth   
]

You are a Strategic Question Generator for complex visual reasoning problems.

Your goal is to generate a \textbf{minimal and sufficient Auxiliary Reasoning Set (ARS)} for a given question. 
The ARS framework decomposes a complex reasoning problem into a sequence of structured sub-questions. 
Each sub-question is directly answerable from the provided information, and together they form a scaffold that enables a language model to solve the original question without ever accessing the image. 
The sub-questions should provide any details that are necessary to answer the original question. 
If there is a sub-question that is not answerable from the provided information, do not include it in the ARS.

\medskip
\textbf{Core Principle}
\begin{itemize}
  \item Directly answerable using the image and original question.
  \item Concrete and visual, avoiding vague or abstract queries.
\end{itemize}

\textbf{Structure of the ARS}

\[
\mathcal{S} = \{(q_1,a_1), (q_2,a_2), \dots, (q_n,a_n)\}
\]

Each sub-question must include:
\begin{itemize}
  \item \texttt{"question"}: A short, visually grounded sub-question.
  \item \texttt{"depends\_on\_sub\_question"}: List of sub-question IDs it depends on (e.g., [Q1, Q2]).
  \item \texttt{"depends\_on\_image"}: yes/no.
  \item \textbf{Most important:} Must not reveal the final answer.
\end{itemize}

\textbf{Required Properties:}
\begin{enumerate}
  \item Sufficiency: Enough to solve the original question.
  \item Minimality: Removing any sub-question makes it unsolvable.
  \item Dependency-restricted: Only reference allowed dependencies.
  \item No redundancy or trivial sub-questions.
  \item Answers must be concise: number, label, or yes/no (prefer numbers/labels).
  \item Do not duplicate the main question.
\end{enumerate}

\textbf{Final Output Format (JSON only):}
\begin{verbatim}
{
  "Q1": {
    "question": "...",
    "depends_on_sub_question": [],
    "depends_on_text": "Yes",
    "depends_on_image": "No"
  },
  ...
}
\end{verbatim}
\end{tcolorbox}
\caption{Full TRACE ARS generation prompt.}
\label{generation_prompt}
\end{figure*}
\end{document}